\documentclass{article}
\usepackage{xcolor,colortbl}

\definecolor{Gray}{gray}{0.85}
\definecolor{LightCyan}{rgb}{0.88,1,1}

\usepackage{ijcai24}

\usepackage{times}
\usepackage{soul}
\usepackage{url}
\usepackage[hidelinks]{hyperref}
\usepackage[utf8]{inputenc}
\usepackage[small]{caption}
\usepackage{graphicx}
\usepackage{amsmath}
\usepackage{amsthm}
\usepackage{booktabs}
\usepackage{algorithm}
\usepackage{algorithmic}
\usepackage[switch]{lineno}

\usepackage{amsmath}
\usepackage{makecell}
\usepackage{multirow}
\usepackage{arydshln}
\usepackage{booktabs}

\usepackage{array}

\title{Improving Zero-Shot Cross-Lingual Transfer via Progressive Code-Switching}

\author{
Zhuoran Li$^1$
\and
Chunming Hu$^{1,2}$\and
Junfan Chen$^{1,2}$\and\\
Zhijun Chen$^1$\and
Xiaohui Guo$^3$\And
Richong Zhang$^{1}$\thanks{$\ \ $Corresponding author}
\affiliations
$^1$SKLSDE, School of Computer Science and Engineering, Beihang University, Beijing, China\\
$^2$School of Software, Beihang University, Beijing, China\\
$^3$Hangzhou Innovation Institute, Beihang University, Hangzhou, China\\
\emails
\{lizhuoranget, hucm, zhijunchen\}@buaa.edu.cn,
\{chenjf, guoxh, zhangrc\}@act.buaa.edu.cn
}

\begin{document}

\maketitle

\begin{abstract}
Code-switching is a data augmentation scheme mixing words from multiple languages into source lingual text. It has achieved considerable generalization performance of cross-lingual transfer tasks by aligning cross-lingual contextual word representations. However, uncontrolled and over-replaced code-switching would augment dirty samples to model training. In other words, the excessive code-switching text samples will negatively hurt the models' cross-lingual transferability. To this end, we propose a \textbf{P}rogressive \textbf{C}ode-\textbf{S}witching (PCS) method to gradually generate moderately difficult code-switching examples for the model to discriminate from easy to hard. The idea is to incorporate progressively the preceding learned multilingual knowledge using easier code-switching data to guide model optimization on succeeding harder code-switching data. Specifically, we first design a difficulty measurer to measure the impact of replacing each word in a sentence based on the word relevance score. Then a code-switcher generates the code-switching data of increasing difficulty via a controllable temperature variable. In addition, a training scheduler decides when to sample harder code-switching data for model training. Experiments show our model achieves state-of-the-art results on three different zero-shot cross-lingual transfer tasks across ten languages. 
\end{abstract}

\section{Introduction}
\begin{figure}[t]
  \centering
  \includegraphics[width=1\linewidth]{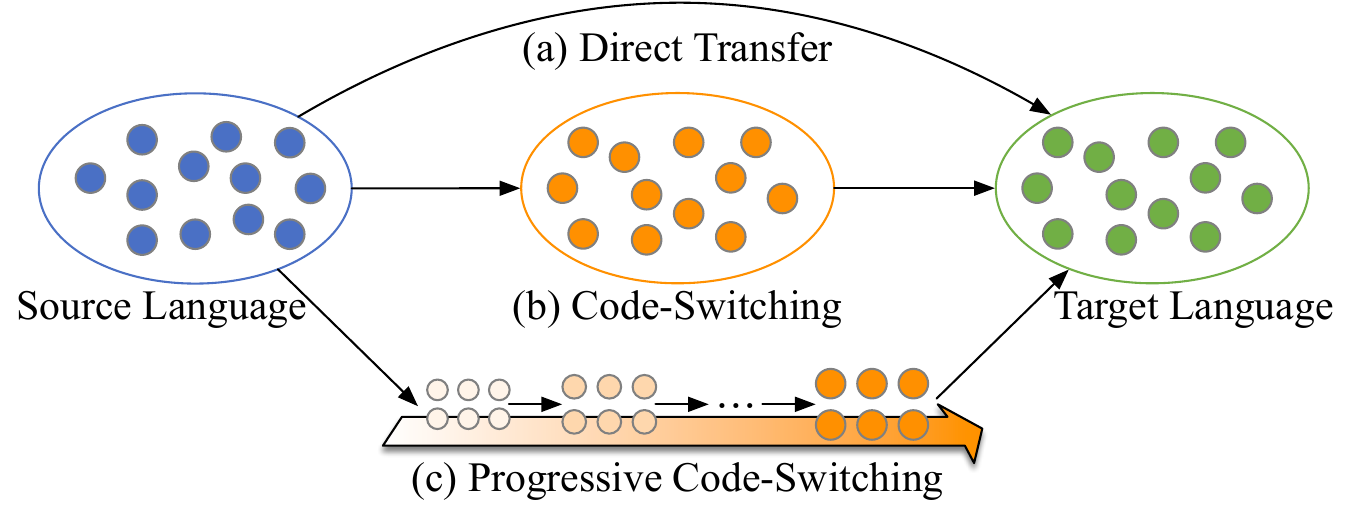}
  \caption{Illustration of our progressive code-switching cross-lingual idea. (a) Direct transfer from source to target. (b) Randomly generating code-switching data. (c) The proposed progressive code-switching method generates code-switching data for the model to discriminate from easy to hard. Larger and darker dots indicate harder code-switching data. }
  \label{fig:intro}
  \hfill
\end{figure}

Zero-shot cross-lingual transfer learning aims to train an adaptable model on a source language that can effectively perform on others without labelled data in the target languages. This study is particularly valuable in circumstances where there are limited or no annotations available for the target languages. In recent years, the multilingual pre-trained language models, such as mBERT~\cite{devlin-etal-2019-bert}, XLM~\cite{conneau-xlm-nips} and XLM-R~\cite{conneau-etal-2020-unsupervised} have achieved significant performance improvements through fine-tuning on source language data and direct application to target language data (as shown in Figure \ref{fig:intro}(a)). Furthermore, it has been discovered that a further multilingual contextualized representation alignment improves the zero-shot cross-lingual transfer performance by exploiting a bilingual dictionary to replace some tokens in the source text with target-lingual translated words. This strategy, named {\it Code-Switching} (CS), usually randomly chooses substitution words and has been shown improvements in many zero-shot cross-lingual tasks~\cite{liu2020attention,qin-cosda-ijcai,zheng-acl-consistency,ma-etal-2022-hcld,10.1145/3580305.3599864} (as shown in Figure \ref{fig:intro}(b)).

Code-Switching, as a data augmentation technique, on one hand, inevitably leads to losing original contextual information when over-replacing words with other lingual synonyms within a sentence. On the other hand, under-replaced code-switching results in insufficient cross-lingual alignment and limited data variation may limit the model's ability to learn and transfer knowledge across languages. Existing studies have indicated that such uncontrolled samples might not necessarily benefit model learning~\cite{qin-cosda-ijcai,yang-etal-2021-multilingual}. 
For example, an original English sentence is ``\textit{All the services were great.}'' and its corresponding multilingual code-switching sentence is ``\textit{todas (ES) les (FR) services waren (DE) great.}'', wherein capital letters in parentheses represent target language abbreviations (e.g. \textit{ES} indicates \textit{Spanish}). Because of very different contextual sentence expressions, such code-switched sentences may not contribute to multilingual word representation alignment, but in some extreme cases, hurt the trained models' cross-lingual generalization ability. Therefore, we should devise some kind of switching scheme to govern the switching extent subtly, e.g., the number of substitutions or the model discrimination ability impact of code-switched sentences.

In this study, we assume that easy code-switching samples could act as pre-training knowledge, which guides the model optimization on harder code-switching data. For instance, we first consider the easy code-switching sentence ``\textit{All les (FR) services were great.}'', which shares substantial contextual overlap with the original sentence. In this case, the model can correctly align the word pair ``\textit{the (EN) - les (FR)}''. Then for a harder code-switching sentence, ``\textit{All les (FR) services waren (DE) great.}'', the previously aligned ``\textit{the (EN) - les (FR)}'' can serve as a pivot for aligning the new word pair ``\textit{were (EN) - waren (DE)}''. By adopting this progressive strategy, new word pairs can be aligned based on the previously identified pairs. Consequently, even the hard code-switching data such as ``\textit{todas (ES) les (FR) services waren (DE) great.}'' would become easier to be handled and thus progressively improving multilingual alignment.

To pursue both effective utilization of code-switching data and model generalization, we borrow the idea of curriculum learning~\cite{bengio2009curriculum} and propose a progressive code-switching framework termed PCS (as shown in Figure \ref{fig:intro}(c)). However, determining the difficulty of code-switching data is challenging, as the importance of each word varies for different tasks. Drawing inspiration from explanation learning methods, we develop a difficulty measurer to estimate the difficulty of code-switching sentences based on the contribution of substitution words toward the prediction. We then introduce a code-switcher with an adjustable temperature parameter to generate appropriate code-switching sentences that align with the current curriculum difficulty. Furthermore, to mitigate the problem of catastrophic forgetting in curriculum learning, we design a scheduler that dynamically adapts the difficulty level to revisit the previously acquired knowledge. 

In summary, we make the following key contributions:
\begin{itemize}
    \item We propose a progressive code-switching method for zero-shot cross-lingual transfer, which mitigates the negative impacts of uncontrolled code-switching data and improves the multilingual representation alignment.
    \item We introduce a word relevance score-guided difficulty measurer, a temperature-adjustable code-switcher, and a dynamic scheduler. They collaboratively regulate the switched samples' difficulty and gradually generate code-switching samples in a controlled manner.
    \item 
    We comprehensively evaluate our proposed approach on three different cross-lingual tasks covering ten different languages. The results demonstrate that PCS substantially enhances performance compared to some strong code-switching baselines.
\end{itemize}

\section{Related Work}

\paragraph{Zero-shot cross-lingual transfer} aims to learn a model with labelled source language data and perform well on other target languages. In recent years, there have been some pre-trained multilingual language models for cross-lingual transfer, such as mBERT~\cite{devlin-etal-2019-bert}, XLM~\cite{conneau-xlm-nips} and XLM-R~\cite{conneau-etal-2020-unsupervised,goyal-etal-2021-larger}. Some studies further improve the alignment of multiple different languages by parallel corpora~\cite{artetxe-schwenk-2019-massively,cao-contextual-alignment,pan-etal-2021-multilingual,chi-etal-2021-infoxlm,iclr-Wei-represent}. Recently, code-switching leverages low-resource bilingual dictionary to align multilingual contextual representations and achieve the state-of-the-art performance in many cross-lingual tasks, such as text classification~\cite{scope10.1145/3459637.3482176}, dialogue system~\cite{liu2020attention,ma-etal-2022-hcld}, sequence tagging tasks~\cite{feng-etal-2022-toward}, and question answering~\cite{DBLP:conf/aaai/NooralahzadehS23}. There are additional attempts to avoid the original signal loss in code-switching~\cite{scope10.1145/3459637.3482176,feng-etal-2022-toward}. Most of the code-switching work in word substitution is random and our work considers the negative impacts of excessive code-switching data. To the best of our knowledge, only a few studies have focused on word selection in code-switching, and none of them investigates progressive code-switching.

\paragraph{Curriculum learning} is proposed as a machine learning strategy by feeding training examples to the model by a meaning order, which is inspired by the learning process of humans and animals~\cite{bengio2009curriculum}. In general, curriculum learning contains a difficulty evaluator used to evaluate the difficulty score of instances, and a scheduler used to decide how examples should be fed to the model. Curriculum learning has been successfully applied to many areas in natural language processing, such as question answering~\cite{sachan-xing-2016-easy}, reading comprehension~\cite{tay-etal-2019-simple}, dialogue system~\cite{shen-feng-2020-cdl,zhu-etal-2021-combining-curriculum-learning} and text classification~\cite{lalor-yu-2020-dynamic,xu-etal-2020-curriculum}. Another line of research aims at providing a theoretical guarantee of curriculum learning, including transfer learning method~\cite{xu-etal-2020-curriculum} and optimization methods~\cite{kumar-nips-self-paced,alex-icml-autocl}ijcai24. In this work, we adopt curriculum learning into code-switching and address the negative impacts of code-switching in cross-lingual transfer.

\section{Progressive Code-Switching}
\begin{figure*}[t]
  \centering
  \includegraphics[width=0.95\linewidth]{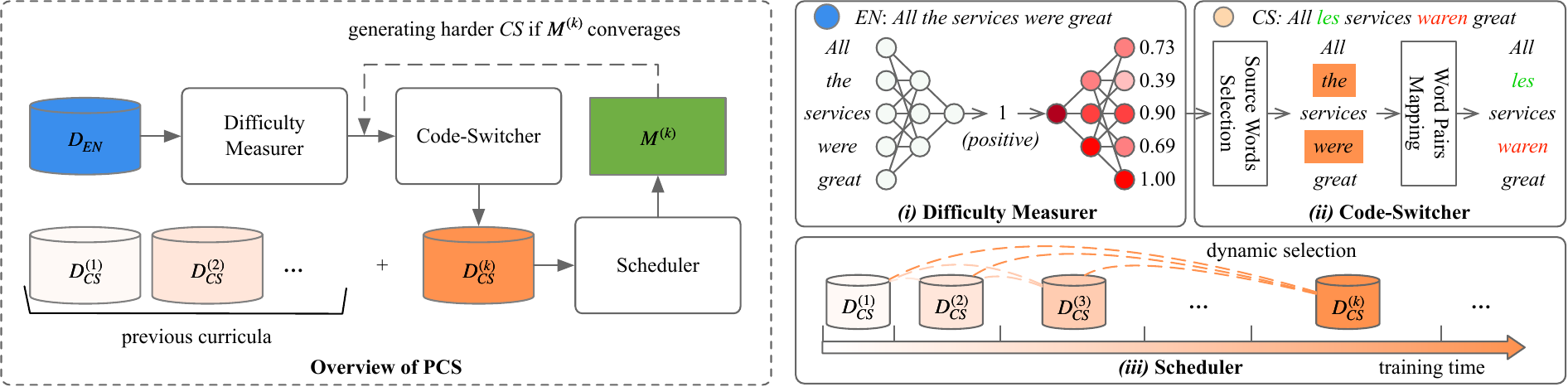}
  \caption{The left subfigure provides an overview of our proposed progressive code-switching framework, while the right subfigure illustrates the three key components. 
  \textbf{(\textit{i})}
  The difficulty measurer calculates the relevance scores to estimate the contribution of each word in the source language data towards the prediction;
  \textbf{(\textit{ii})}
  The code-switcher selects substitution words based on the relevance score to generate suitable code-switching data;
   \textbf{(\textit{iii})}
  The scheduler decides when to sample harder code-switching examples for model training. $D_{EN}$: the labelled data in the source language; $D^{(k)}_{CS}$: the generated code-switching data in the $k$-th curriculum; $M^{(k)}$: the learned model for target languages in the $k$-th curriculum. 
}
  \label{fig:model}
\end{figure*}

We introduce \textbf{P}rogressive \textbf{C}ode-\textbf{S}witching (PCS) method in this section, which can be applied to various zero-shot cross-lingual transfer learning downstream tasks. Figure \ref{fig:model} depicts an overview of the PCS framework. We will describe the details of our approach from the following four components: difficulty measurer, code-switcher, scheduler, and model trainer.

\paragraph{Problem Formulation.}
Formally in a zero-shot cross-lingual transfer task, given a source language sentence ${\mathbf{x}}=\{x_1, x_2, ..., x_L\}$ with $L$ words, our code-switcher module generates the augmented code-switching sentence ${{\mathbf{x}}^a}=\{x_1^a, x_2^a, ..., x_L^a\}$ by replacing or not $x_i$ with $x_i^a$ from the predefined bilingual dictionary according to word relevance and difficulty temperature. The label $\mathbf{y}$ is kept the same as the original sentence. With the word relevance score of each word $x_i$ measured in a sentence, the code-switcher gradually generates harder code-switching sentences. A pre-trained difficulty measurer with the source language data determines the word relevance. Our goal is to learn a model with source language data ${D}_{S}^{train}=\{({\mathbf{x}}, \mathbf{y})\}$ and augmented code-switching data ${D}_{CS}^{train}=\{({{\mathbf{x}}^a}, \mathbf{y})\}$ to perform zero-shot prediction on target languages ${D}_{T}^{test}=\{{\mathbf{x}}\}$. 

\subsection{Difficulty Measurer}

The idea of our PCS lies in the strategy of “training from easier code-switching data to harder code-switching data”. Due to the code-switching data being augmented from the original data without a predefined difficulty score, we first need to measure what kinds of data are harder than others. Existing popular difficulty measurers in natural language tasks include sentence length~\cite{spitkovsky-etal-2010-baby},  word rarity~\cite{platanios-etal-2019-competence}, and replacement ratio~\cite{wei-etal-2021-shot}. However, these measures ignore that replacing the same words produces different degrees of change under different tasks. Therefore, this cannot guarantee the `easy-to-hard' order, because a code-switching sentence with a few important words being replaced is more difficult than one with a large of irrelevant words being replaced for the model. Here, we argue that code-switching sentences having more important words being replaced have bigger semantic distortion with the original sentences than others, and are treated as more difficult instances. Inspired by explanation methods~\cite{arras-etal-2019-evaluating}, we use Layer-Wise Relevance Propagation (LRP)~\cite{bach2015pixel} to assign each word a relevance score indicating to which extent it contributed to a particular prediction, as the basis for estimating the difficulty level of CS. In other words, LRP can quantify whether a
token is important in the model’s decisions to the prediction we are interested in.

We suppose given a trained model $f$, which has learned a scalar-valued prediction function, e.g. $f_c(\mathbf{x})$ means prediction probability of an input sequence $\mathbf{x}$ being class $c$ in the classification task. 
We adopt a BERT-based LRP~\cite{wu2021explaining} consisting of a standard forward pass, followed by a specific backward pass. For a linear layer of the form as Eq. (\ref{eq:z_j}), and given the relevance of the output neurons $r_j$, the relevance of the input neurons $r_i$ are computed through the following Eq. (\ref{eq:R_i}), where $\epsilon$ is a stabilizer. 
\begin{equation}
\label{eq:z_j}
    {z}_{j}^{(l+1)} = \sum_{i}{{x}_{i}^{(l)} \cdot {w}_{ij}^{(l)}+{b}_{j}^{(l+1)}}
\end{equation}

\begin{equation}
\label{eq:R_i}
    {r}_{i}^{(l)} = \sum_{j} \frac{{{x}_{i}^{(l)} \cdot {w}_{ij}^{(l)}}}{z_j^{(l+1)}+\epsilon} \cdot {r}_{j}^{(l+1)}
\end{equation}

In practice, starting from the output neuron whose relevance is set to the value of the prediction function, i.e. $f_c(\mathbf{x})$, LRP uses Eq. (\ref{eq:R_i}) to iteratively redistribute the relevance from the last layer $f_c(\mathbf{x})$ down to the input layer $\mathbf{x}$, layer by layer, and verifies a relevance conservation property. We denote ${r}_{(d)}(x)$ the relevance of the $d$-th dimension ($d \in\{1,2,...,D\}$) of the token $x$ and we can derive it as follow Eq. (\ref{eq:R_k}):

\begin{align}
\label{eq:R_k}
{r}_{(d)}(x) &= f_{c}(\mathbf{x})(\frac{\mathbf{w}^{(l)}{x}^{(l)}}{\mathbf{z}^{(l+1)}})a'({z}^{(l+1)}_{j})...(\frac{\mathbf{w}^{(0)}{x}^{(0)}}{\mathbf{z}^{(1)}})a'({z}^{(1)}_{j})  \notag  \\ 
&= f_{c}(\mathbf{x})(\prod_{l}{\frac{\mathbf{w}^{(l)}{x}^{(l)}}{\mathbf{z}^{(l+1)}}})(\prod_{l}{a'({z}^{(l+1)}_{j})})  \notag  \\
&\approx f_{c}(\mathbf{x})(\prod_{l}{\frac{\mathbf{z}^{(l)}}{\mathbf{z}^{(l+1)}}})
\end{align}

where $\mathbf{z}^{(l)}$ is column matrix of hidden states in layer $l$, and derivatives of non-linear activation functions $a'(\cdot)$ are ignored as proposed in~\cite{arras-etal-2019-evaluating,wu2021explaining}. For non-linear layers such as the self-attention layer and the residual layer, ${z}^{l}$ is approximated by the first term in the Taylor expansion formally as Eq. (\ref{eq:z^l}) as proved in~\cite{bach2015pixel}, where $f_{\psi}$ is an arbitrary differentiable function, and $\hat{{x}}$ is the Taylor base point where $f_{\psi}({\hat{{x}}})=0$. We derive the relevance score of the token ${r}(x)$ w.r.t. the class $c$ using absolute sum of ${r}_{(d)}(x)$, i.e. ${r}(x)=\sum_{d}{{r}_{(d)}(x)}$. 

\begin{equation}
\label{eq:z^l}
    {z}_j^{(l+1)} = \sum_{i} f_{\psi}({x}_i^{(l)}) \approx \sum_{i}  \frac{\partial f_{\psi}(\hat{{x}}_{i}^{(l)})}{\partial {x}_{i}^{(l)}} ({x}_{i}^{(l)}-\hat{{x}}_{i}^{(l)})
\end{equation}

Via this backward pass, we can observe which words really contributed to the output. An example is shown in Figure \ref{fig:model}, ``\textit{services}'' and ``\textit{great}'' significantly contribute to the prediction of ``\textit{positive}''. 

\subsection{Code-Switcher}
To generate code-switching sentences that match the difficulty level of the current curriculum, we introduce a code-switcher that incorporates a variable temperature denoted as $\tau$. This temperature parameter represents the proportion of words to be replaced, and it increases linearly as the curriculum stage advances. Given the original source-language sentence and the word relevance score, the code-switcher selects the words in ascending order of word relevance. After that, a target language is chosen randomly based on a bilingual dictionary. It's important to note that source-language words can have multiple translations in the target language. In this case, one of the multiple translations is randomly selected as the target word. While this selection might not guarantee an exact word-to-word translation within the context, we have observed that this scenario is infrequent within most bilingual dictionaries. Consequently, the randomness introduced by this process has a minimal impact on our code-switching.

\subsection{Scheduler}
The scheduler aims to sample the data and send it to the model trainer for training. The scheduler decides when to sample the harder training data with the training progress. For our PCS, we begin with a temperature of $\tau=0$ equivalent to sampling the source language data. Then, the temperature linearly increases by the increment $\delta$ (e.g. $\delta=0.1$) every time the validation loss convergence, up to a final temperature of $\tau=1$. As the temperature increases, we will generate harder code-switching data. To encourage the model to pay more attention to harder data, we set larger early stopping patience for harder curricula than easier ones. However, we found that training the model on a sequence of CS datasets faces the problem of catastrophic forgetting~\cite{kirkpatrick2017overcoming}. As the curriculum stage progresses, code-switching training datasets with varying augmentation levels are sequentially inputted into the model. This results in the modification of weights acquired during the initial curriculum once the model encounters the target of the new curriculum, causing the occurrence of catastrophic forgetting. To mitigate this problem, we design a dynamic curriculum scheduler for the model to revisit previous curricula. Specifically, at the $k$-th curriculum stage, the scheduler selects the code-switching data $D^{(i)}_{CS}$ for the model training on the following probability:
\begin{equation}
\label{eq:p_CS}
    {P}(D^{(i)}_{CS}) = \frac{e^{i-k}}{\sum_{i=1}^{k}{e^{i-k}}}
\end{equation}

\subsection{Model Trainer}
The model trainer progressively trains downstream task-specific models with the training data given by the scheduler. And the model has the same network architecture as the pre-trained model in difficulty measurer. We use the conventional fine-tuning method as proposed in the~\cite{devlin-etal-2019-bert}. Specifically, we use a pre-trained multilingual model as an encoder to obtain the representation. Then model $f$ predicts task-specific probability distributions, and we define the loss of cross-lingual fine-tuning as 
\begin{equation}
\label{eq:trainer}
    \mathcal{L}_{task} = -\sum_{{x}} {l}(f({x}), G({x}))
\end{equation}
where $G(\mathbf{x})$ denotes the ground-truth label of example $\mathbf{x}$, ${l}(\cdot, \cdot)$ is the loss function depending on the downstream task. 

\begin{table}[t]
\centering
\scalebox{1}{
\setlength{\tabcolsep}{0.8mm}{
\begin{tabular}{lcccccc}
\toprule
\textbf{Dataset} & \textbf{\#Lang.} & \textbf{\#Train} & \textbf{\#Dev.} & \textbf{\#Test} & \textbf{\#Labels} & \textbf{Metric} \\
\midrule
{PAWS-X} & {7} & {49,401} & {2,000} & {2,000} & {2} & {Acc.} \\
{MLDoc} & {8} & {10,000} & {1,000} & {2,000} & {4} & {Acc.} \\
{XTOD} & {3} & {30,521} & {4,181} & 2,368 & {12/11} & {Acc./F1}  \\
\bottomrule
\end{tabular}
}
}
\caption{Summary statistics of datasets. Note that XTOD is a joint task dataset that includes 12 intent labels and 11 slot labels.
}
\label{exp:dataset_statistics}
\end{table}

\section{Experiments}
\subsection{Setup}
\paragraph{Tasks and Datasets.} 
To comprehensively evaluate our proposed method, we conduct experiments on three types of cross-lingual transfer tasks with three widely used datasets.  
(1) For {paraphrase identification},  we employ \textbf{PAWS-X} dataset~\cite{yang-etal-2019-paws} containing seven languages. The label has two possible values: 0 indicates the pair has a different meaning, while 1 indicates the pair is a paraphrase. The evaluation is the classification accuracy (ACC). (2) For {document classification}, we employ \textbf{MLDoc}~\cite{SCHWENK18.658} as our document classification dataset, including seven different target languages. The evaluation is the classification accuracy (ACC). (3) For {spoken language understanding}, we use the cross-lingual task-oriented dialogue dataset (\textbf{XTOD})~\cite{schuster-etal-2019-cross-lingual} including English, Spanish, and Thai across three domains. The corpus includes 12 intent types and 11 slot types, and the model has to detect the intent of the user utterance and conduct slot filling for each word of the utterance. The performance of intent detection is evaluated using classification accuracy (ACC), while slot filling can be stated as a sequence labelling task evaluated on the F1 score. The statistics of datasets are summarized in Table \ref{exp:dataset_statistics}.

\begin{table}[t]
\centering
\setlength{\tabcolsep}{0.9mm}{
\scalebox{1}{
\begin{tabular}{l c c c c c c >{\columncolor{lightgray!20}}c}
\toprule
\textbf{Model} & {\textbf{de}} & {\textbf{es}} & {\textbf{fr}} & {\textbf{ja}} & {\textbf{ko}} & {\textbf{zh}} & \textbf{\underline{Avg.}}\\
\midrule
\multicolumn{8}{l}{mBERT-based models}  \\
\midrule
mBERT~[\citeyear{devlin-etal-2019-bert}] & 85.7 & 87.4 & 87.0 & 73.0 & 69.6 & 77.0 & 80.0 \\
{WS}~[\citeyear{scope10.1145/3459637.3482176}] & 86.7 & 89.8 & 89.4 & 78.9 & 78.1 & 81.7 & 84.1 \\
{SCOPA}~[\citeyear{scope10.1145/3459637.3482176}] & 88.7 & 90.3 & 89.7 & \textbf{81.5} & 80.1 & 84.3 & 85.8 \\
{SALT}~[\citeyear{wang2023selfaugmentation}] & 87.9 & 89.9 & 89.1 & 78.6 & 77.4 & 81.8 & 84.1 \\
IECC~[\citeyear{ji-etal-2023-isotropic}] & 87.9 & 88.9 & 89.3 & 79.4 & 77.9 & 81.8 & 84.2 \\
Macular~[\citeyear{10.1145/3580305.3599864}] & 88.1 & 90.0 & 89.3 & 80.3 & 79.0 & 83.6 & 85.1 \\
\textbf{PCS (Ours)} & \textbf{89.5} & \textbf{91.4} & \textbf{90.9} & 80.8 & \textbf{80.4} & \textbf{84.6} & \textbf{86.3} \\
\midrule
\multicolumn{8}{l}{larger XLM-R-based models}  \\
\midrule
{XLM-R}~[\citeyear{conneau-etal-2020-unsupervised}]  & 89.7 & 90.1 & 90.4 & 78.7 & 79.0 & 82.3 & 85.0 \\
TCS~[\citeyear{lu-etal-2023-take}] & 90.8 & 91.6 & 91.4 & 81.8 & 81.7 & 84.7 & 87.0 \\
SCS~[\citeyear{lu-etal-2023-take}] & 91.7 & 91.6 & 92.0 & 82.8 & 82.9 & 85.3 & 87.7 \\
IECC~[\citeyear{ji-etal-2023-isotropic}] & 92.0 & 92.1 & 92.6 & \textbf{83.7} & 84.3 & {85.6} & 88.4 \\
\textbf{PCS (Ours)}  & \textbf{92.4} & \textbf{93.0} & \textbf{92.8} & {83.6} & \textbf{85.1} & \textbf{86.6} & \textbf{88.9} \\
\bottomrule
\end{tabular}
}
}
\caption{
Results (Acc.) on natural language inference (PAWS-X). The last `\textbf{\underline{Avg.}}' column denotes the average result for all languages. The best performance is in \textbf{bold} (same for Tables \ref{exp:main-mldoc} and \ref{exp:main-xtod}).
}
\label{exp:main-paws-x}
\end{table}

\begin{table}[t]
\centering
\setlength{\tabcolsep}{0.75mm}{
\scalebox{1}{
\begin{tabular}{l ccccccc   >{\columncolor{lightgray!20}}c}
\toprule
\textbf{Model}  & \textbf{de} & \textbf{es} & \textbf{fr} & \textbf{it} & \textbf{ja} & \textbf{ru} & \textbf{zh} & \textbf{\underline{Avg.}}\\
\midrule
\multicolumn{8}{l}{mBERT-based models}  \\
\midrule
mBERT~[\citeyear{devlin-etal-2019-bert}] & 80.2 & 72.6 & 72.6 & 68.9 & 56.5 & 73.7 & 76.9 & 71.6\\
{CoSDA}~[\citeyear{qin-cosda-ijcai}] & 86.3 & 79.2 & 86.7 & 72.6 & 73.7 & 75.1 & {85.5} & 79.9\\
{WS}~[\citeyear{scope10.1145/3459637.3482176}] & {89.1} & {76.7} &  88.1 & 72.0 & 74.4 & 79.0 & {83.0} & 80.3\\
{SCOPA}~[\citeyear{scope10.1145/3459637.3482176}] & {90.7} & {86.1} &  \textbf{90.5} & {75.1} & 76.7 & \textbf{80.4} & {85.5} & 83.6\\
{M-BoE}~[\citeyear{nishikawa-etal-2022-multilingual}] & {75.5} & {76.9} &  {84.0} & {70.0} & 71.1 & {68.9} & {72.2} & 74.1 \\
\textbf{PCS (Ours)}  & \textbf{91.4} & \textbf{87.6} & \textbf{90.5} & \textbf{78.4} & \textbf{78.3} & {78.1}  & \textbf{88.9} & \textbf{84.7}\\
\midrule
\multicolumn{8}{l}{larger XLM-R-based models}  \\
\midrule
{XLM-R} & 94.9 & 94.5 & 94.7 & \textbf{85.6} & 81.9 & 72.0 & 91.8 & 87.9 \\
\textbf{PCS (Ours)} & \textbf{95.4} & \textbf{95.1} & \textbf{95.6} & \textbf{85.6} & \textbf{83.2} & \textbf{72.5} & \textbf{92.6} & \textbf{88.6} \\
\bottomrule
\end{tabular}
}
}
\caption{
Results (F1) on document classification (MLDoc). 
}
\label{exp:main-mldoc}
\end{table}

\begin{table}[t]
\centering
\scalebox{1}{
\begin{tabular}{l c c   >{\columncolor{lightgray!20}}c}
\toprule
\textbf{Model} & {\textbf{es}} & \textbf{th} & \textbf{\underline{Avg.}}\\
\midrule
\multicolumn{4}{l}{mBERT-based models}  \\
\midrule
mBERT~[\citeyear{devlin-etal-2019-bert}] & 73.7/51.7 & 28.2/10.6 & 51.0/31.2 \\
{MLT}~[\citeyear{liu2020attention}] & 86.5/74.4 & 70.6/28.5 & 78.6/51.5 \\
{CoSDA}~[\citeyear{qin-cosda-ijcai}] & 94.8/80.4 & 76.8/37.3 & 85.8/58.9 \\
{HCLD}~[\citeyear{ma-etal-2022-hcld}] & {84.7}/{79.5} & {\textbf{81.0}}/{32.2}  & 82.9/55.9 \\
\textbf{PCS (Ours)} & \textbf{95.3}/\textbf{81.5} & \textbf{81.0}/\textbf{38.5} & \textbf{88.2}/\textbf{60.0} \\
\midrule
\multicolumn{4}{l}{larger XLM-R-based models}  \\
\midrule
{XLM-R} & 96.8/85.5 & 95.4/32.8 & 96.1/59.2 \\
\textbf{PCS (Ours)} & \textbf{98.0}/\textbf{86.6} & \textbf{97.0}/\textbf{54.3} & \textbf{97.5}/\textbf{70.4} \\
\bottomrule
\end{tabular}
}
\caption{
Results (Intent Acc./Slot F1) on slot filling and intent detection (XTOD).
}
\label{exp:main-xtod}
\end{table}

\paragraph{Implementation Details.} We implement our proposed method based on mBERT and XLM-R-large of HuggingFace Transformer \footnote{\url{https://github.com/huggingface/transformers}.} as the backbone model. We set our hyperparameters empirically following previous approaches~\cite{liu2020attention,qin-cosda-ijcai,scope10.1145/3459637.3482176} with some modifications. We set the batch size to 16 or 64, the maximum sequence length to 128, and the dropout rate to 0.1, and we use AdamW as the optimizer. We select the best learning rate from \{5e-6, 1e-5\} for the encoder and \{1e-3, 1e-5\} for the task-specific network layer. As for the scheduler, we initialize $\tau=0$, which linearly increases as the stage increases. We conduct each experiment 3 times with different random seeds and report the average results of 3 run experiments. To imitate the zero-shot cross-lingual setting, we consider English as the source language and others as the target languages. The bilingual dictionaries we use for code-switching are from MUSE~\cite{lample-muse}.  All models are trained on a single Tesla V100 32GB GPU. 

\subsection{Performance Comparison}

We compare our method with the following competitive code-switching enhancement models as our baselines:

\paragraph{mBERT}~\cite{devlin-etal-2019-bert} is a 12-layer transformer model pre-trained on the Wikipedias of 104 languages and is fine-tuned only on the labelled source language training data. 

\paragraph{XLM-R}~\cite{conneau-etal-2020-unsupervised} is a 24-layer transformer-based multilingual masked language model pre-trained on a text in 100 languages around 2.5TB unlabeled text data extracted from CommonCrawl datasets. 

\paragraph{MLT}~\cite{liu2020attention} chooses source keywords based on the attention scores computed by a trained source language task-related model to generate code-switching sentences.

\paragraph{CoSDA}~\cite{qin-cosda-ijcai} generates code-switching data with an empirically constant token replacement ratio to enhance the multilingual representations.

\paragraph{WS}~\cite{scope10.1145/3459637.3482176} simply substitutes words in sentences in every batch during training. 

\paragraph{SCOPA}~\cite{scope10.1145/3459637.3482176} softly mixups the source word embeddings and the switched target word embeddings with an auxiliary pairwise alignment objective. 

\paragraph{M-BoE}~\cite{nishikawa-etal-2022-multilingual} only mixups the embeddings of Wikipedia entities to boost the performance of cross-lingual text classification.

\paragraph{HCLD}~\cite{ma-etal-2022-hcld} classifies pre-defined intent with code-switching augmentation and then fills the slots under the guidance of intent. 

\paragraph{SALT}~\cite{wang2023selfaugmentation} incorporates masked language modelling-based offline code-switching and online embedding mixup to enhance the cross-lingual transferability.

\paragraph{IECC}~\cite{ji-etal-2023-isotropic} proposes an isotropy enhancement and constrained code-switching method for cross-lingual transfer to alleviate the problem of misalignment.

\paragraph{TCS}~\cite{lu-etal-2023-take} encourages cross-lingual interactions via performing token-level code-switched masked language modelling.

\paragraph{SCS}~\cite{lu-etal-2023-take} further proposes a semantic-level code-switched masked language modelling based on multiple semantically similar switched tokens in different languages.

\paragraph{Macular}~\cite{10.1145/3580305.3599864} incorporates code-switching augmentation into multi-task learning to capture the common knowledge across tasks and languages.

\begin{figure*}[t]
  \centering
  \includegraphics[width=1\linewidth]{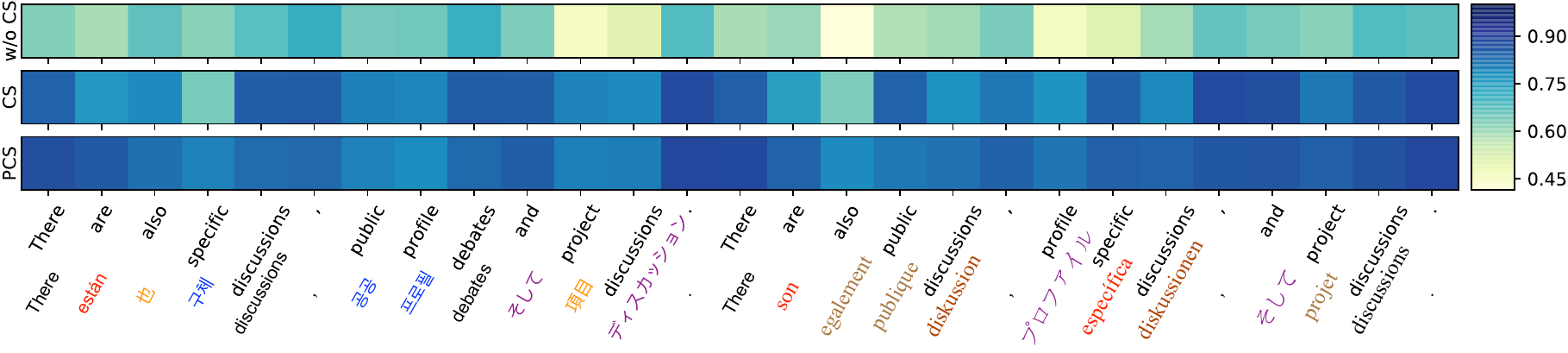}
  \caption{A darker colour indicates a higher cosine similarity score between source words in the original sentence and corresponding target words in the code-switching sentence.} 
  \label{exp:case_sim_score}
  \hfill  
\end{figure*}

As shown in Tables \ref{exp:main-paws-x}, \ref{exp:main-mldoc} and \ref{exp:main-xtod}, compared with strong code-switching baselines, PCS shows its superiority and generality across different backbones and tasks at the zero-shot setting. In MLDoc and XTOD, we implement the XLM-R baseline following the reported settings of XLM-R~\cite{conneau-etal-2020-unsupervised}. In Table \ref{exp:main-paws-x}, PCS outperforms SCOPA by 0.5\% based on mBERT, and outperforms IECC by 0.5\% based on XLM-R. For Table \ref{exp:main-mldoc}, PCS outperforms SCOPA by 1.1\% based on mBERT, and outperforms our reproduced XLM-R by 0.7\%. In Table \ref{exp:main-xtod}, compared to random selection (CoSDA) or solely choosing keywords (MLT) to construct code-switching sentences, our approach demonstrates superior performance.

\begin{table}[t]
\small
\centering
\setlength{\tabcolsep}{1mm}{
\scalebox{1}{
\begin{tabular}{l c c c c c c >{\columncolor{lightgray!20}}c}
\toprule
\textbf{Model} & {\textbf{de}} & {\textbf{es}} & {\textbf{fr}} & {\textbf{ja}} & {\textbf{ko}} & {\textbf{zh}} & \textbf{\underline{Avg.}}\\
\midrule
\textbf{PCS (Full)} & \textbf{89.5} & \textbf{91.4} & {90.9} & \textbf{80.8} & \textbf{80.4} & \textbf{84.6} & \textbf{86.3} \\
(1) {w/o scheduler} & 88.7 & 91.2 & 90.5 & 80.1 & 80.1 & 83.5 & 85.7  \\
(2) {w/o CL} & 87.7 & 90.1 & 89.9 & 79.2 & 78.9 & 82.8 & 84.8 \\
(3) {using Ratio-CL} & 88.5 & 90.7 & 90.7 & 79.3 & 79.8 & 83.2 & 85.4 \\
(4) {using Grad-CL} & 89.4 & 91.0 & \textbf{91.1} & 80.2 & 80.3 & 84.1 & 86.0 \\
(5) {using Anti-CL} & 88.1 & 91.0 & 90.5 & 80.5 & 79.6 & 84.4 & 85.7 \\
(6) {using TGT-Only} & 89.0 & 90.4 & 90.0 & 79.4 & 79.5 & 82.9 & 85.2 \\
\bottomrule
\end{tabular}
}
}
\caption{
Ablation study (Acc.) on PAWS-X.
}
\label{exp:ablation-paws-x}
\end{table}

\begin{table*}[htpb]
\centering
\begin{tabular}{lcc}
\toprule
\textbf{Model} & \textbf{Sentence Pair} & \textbf{Prediction}\\
\midrule
\makecell*[c]{w/o CS} & 
\makecell*[l]{
Un A Khap es un clan o grupo de clanes relacionados, principalmente de los jats del \colorbox{green!40}{oeste}
de\\ Uttar Pradesh y del \colorbox{green!30}{este} de \colorbox{green!15}{Haryana}. Un khap es un clan, o grupo de clanes relacionados,\\ principalmente entre los jats del \colorbox{green!40}{este} de Uttar Pradesh y el \colorbox{green!30}{oeste} de Haryana. 
} & same \\
\makecell*[c]{CS} & 
\makecell*[l]{
Un A \colorbox{green!15}{Khap} es un clan o grupo de clanes \colorbox{green!40}{relacionados,} principalmente de los jats del
oeste de\\ Uttar Pradesh y del este de \colorbox{green!30}{Haryana.} Un \colorbox{green!40}{khap} es un clan, o grupo de clanes \colorbox{green!30}{relacionados,}\\ principalmente entre los jats del este de Uttar Pradesh y el oeste de Haryana.  
} & same \\
\makecell*[c]{PCS} & 
\makecell*[l]{
Un \colorbox{green!15}{A} Khap es un clan o grupo de clanes relacionados, principalmente de los jats del oeste de\\ Uttar Pradesh y del \colorbox{green!30}{este} de \colorbox{green!40}{Haryana.} Un khap es un clan, o grupo de
clanes relacionados,\\ principalmente entre los jats del este de Uttar Pradesh y el \colorbox{green!30}{oeste} de \colorbox{green!40}{Haryana.} 
} & different \\
\bottomrule
\end{tabular}
\caption{Case study on a Spanish pair having different (the golden label) semantic meaning in paraphrase identification task. The green-highlighted words represent the top five words that contribute the most to the prediction.}
\label{exp:case_keywords}
\end{table*}

\subsection{Ablation Study} 
To better understand PCS, we conduct ablation studies to analyse the contributions of each component. Table \ref{exp:ablation-paws-x} presents the ablation study results for our PCS on PAWS-X. Comparing the full model, we can draw several conclusions: (1) We remove our dynamic scheduler in PCS for the variant {w/o scheduler}. Results show that our dynamic curriculum selection effectively alleviates the problem of catastrophic forgetting. (2) We remove the curriculum learning strategy for the variant {w/o CL}, which degrades into the random code-switching model. Results demonstrate that the usage of PCS pushes code-switching by an absolute gain of 1.5\% on average. For (3), we use the word replacement ratio as the difficulty measure in PCS. Results show that the performance drops about 0.9\% on average because the word replacement ratio cannot flexibly measure the difficulty of a code-switching sentence for different tasks. For (4), as an alternative difficulty measurer in our study, we employ a gradient-based explanation model. Results show that the performance slightly drops about 0.3\% on average. This means our LRP-based difficulty measure is superior to the gradient-based method. For (5), we test anti-curriculum learning, which progressively generates code-switching data from hard to easy. Results show that the performance drops about 0.6\% on average. This indicates that introducing hard data in the early curriculum negatively impacts learning performance. For (6), we generate code-switching sentences that only mix one target language with the source language. The performance of each language is more degraded than the PCS. This indicates that mixing multiple languages in code-switching can enhance the model's cross-lingual transfer ability.

\begin{figure}[t]
  \centering
  \includegraphics[width=0.9\linewidth]{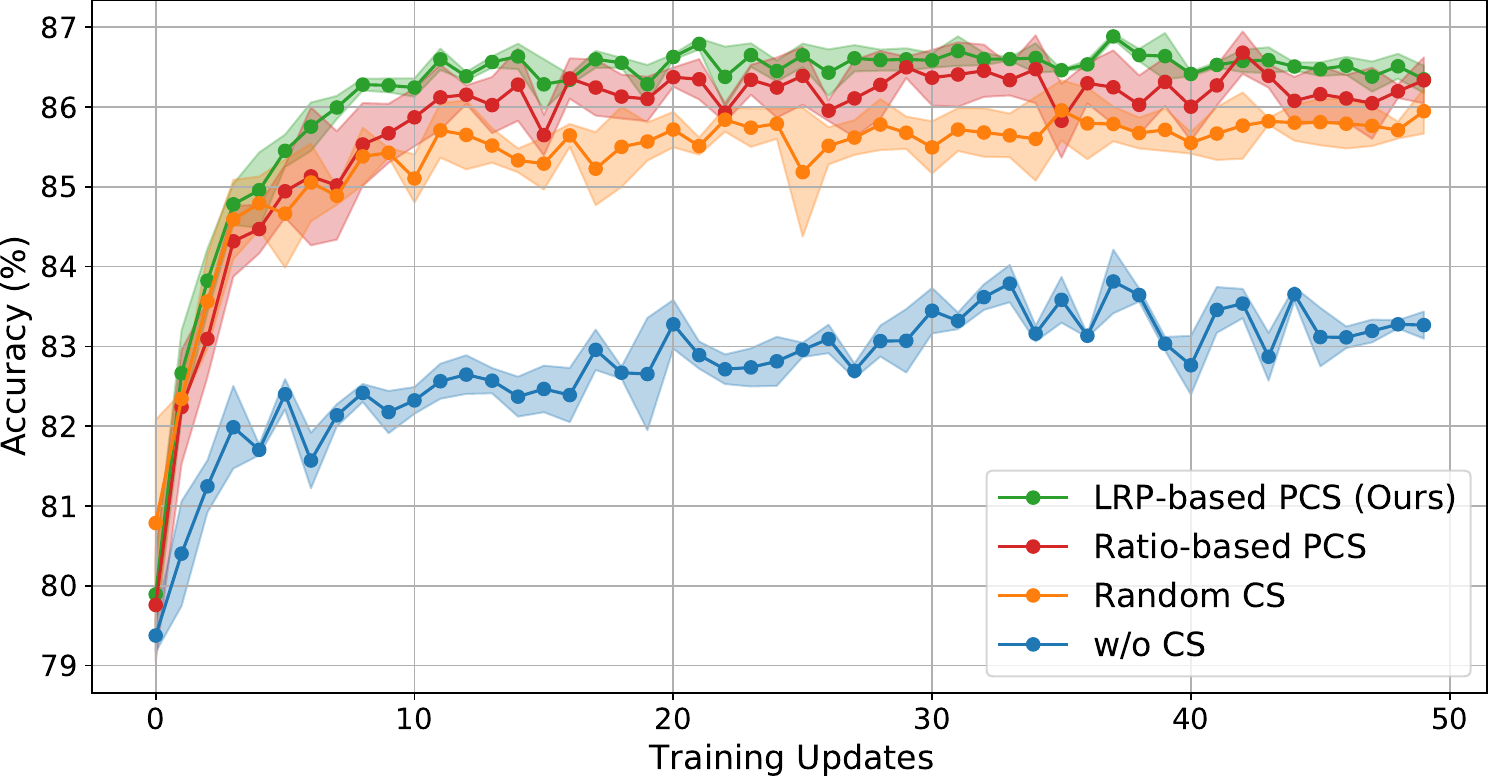}
  \caption{Learning curves of our PCS and three baseline models on PAWS-X based on mBERT.}
  \label{fig:learning_curve}
  \hfill
\end{figure}

\subsection{Case Study}
By analyzing the typical cases bettered by PCS, we seek to shed light on the underlying reasons behind its success.

Firstly, PCS can improve the quality of multilingual word representations. Specifically, by leveraging the high-quality word alignment obtained from easier code-switching sentences and their original counterparts, the model gains valuable pivots to comprehend words within over-replaced code-switching sentences. As illustrated in Figure \ref{exp:case_sim_score}, the representations of ``specific'' and ``also'' exhibit a relatively low similarity score with their corresponding code-switched words in the first two rows. This is because the model without code-switching (w/o CS) and the code-switching model (CS) simultaneously consider all words within the over-replaced code-switching sentence. On the other hand, our Progressive Code-Switching (PCS) provides a higher similarity score for these word pairs. This is because PCS incorporates other word pairs aligned in the early curricula, allowing it to understand the over-replaced code-switching sentence. 

Moreover, PCS helps the model focus on task-relevant keywords, enabling accurate predictions. We calculate the relevance scores of each word for the prediction result, and we notice that PCS makes correct predictions and provides understandable justifications. As illustrated in Table \ref{exp:case_keywords}, PCS correctly predicts that the sentence pair is semantically different by focusing on the discriminative words (\textit{``este''}, \textit{``Haryana''}, \textit{``oeste''} and \textit{``Haryana''}). In contrast, the model without code-switching (w/o CS) and the code-switching model (CS) fail to do so. We conjecture that our PCS tends to better understand task-related keywords due to its learning process in the later stages of the curriculum, during which the model has already acquired some multilingual knowledge. This further confirms the effectiveness and generalization of the proposed PCS for different tasks.

\subsection{Learning Curve Analysis} 
To assess the effectiveness of the progressive code-switching method, we compare the learning curves of our LRP-based PCS with the ratio-based progressive code-switching, the random code-switching and the vanilla fine-tuned mBERT (without code-switching). In Figure \ref{fig:learning_curve}, we draw the average accuracy and the standard deviation of all target languages based on three experimental runs during the model training updates on PAWS-X. Firstly, all three code-switching models demonstrate significantly improved performance compared to the vanilla mBERT baseline. This indicates that introducing code-switching enhances the model's cross-lingual capabilities. Secondly, both progressive CS models converge to a better solution than the traditional random CS model using the same number of updates. This means that progressive code-switching facilitates the model to converge rapidly to a better minimum. Thirdly, the ratio-based PCS exhibits instability at 15-th and 35-th updates, unlike our LRP-based PCS. This instability arises from the sub-optimal static word replacement rate difficulty measurer, which cannot guarantee that the generated code-switching data satisfies the nature from easy to difficult for various tasks. In contrast, our LRP-based model solves this problem by dynamically generating code-switching data guided by the token-level relevance scores derived from the trained task-related model.

\subsection{Multilingual Representation Visualization}
\begin{figure}[!t]
\centering
\begin{tabular}{cc}
    \hspace{-0.35cm} \includegraphics[width=0.49\linewidth]{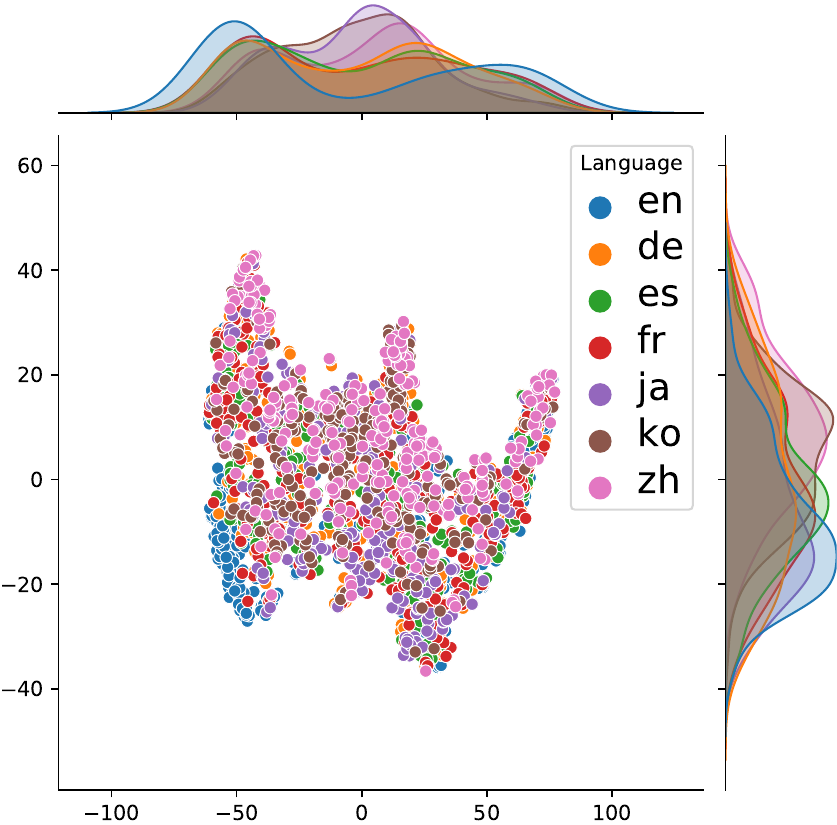} &
    \hspace{-0.35cm} \includegraphics[width=0.49\linewidth]{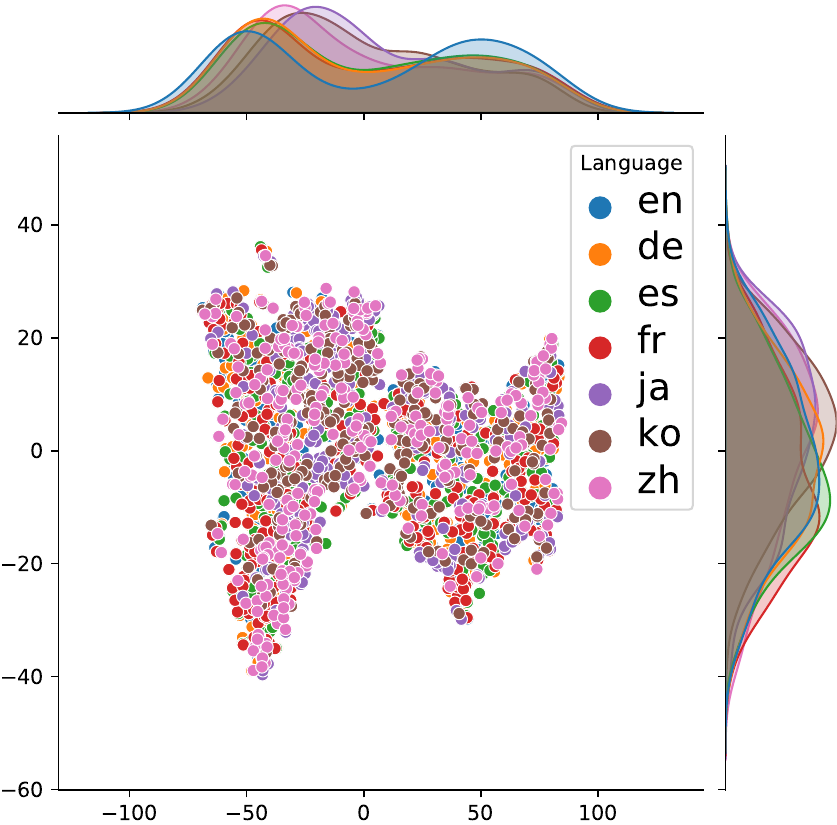} \\
    \hspace{-0.35cm} (a) mBERT & 
    \hspace{-0.35cm} (b) Our PCS 
\end{tabular}
\caption{Multilingual alignment t-SNE visualization. Sentence embeddings from fine-tuned mBERT and our PCS.}
\label{exp:tSNE}
\end{figure}

To examine the alignment of multiple languages, we compare the alignment results of vanilla fine-tuned mBERT and our PCS in terms of sentence representation. We select sentences represented by seven different languages from the PAWS-X datasets respectively to visualize the embedding space. As shown in Figure \ref{exp:tSNE} (a), different languages are distributed in different positions in the embedding space, which indicates that fine-tuned mBERT can distinguish them easily. In contrast, Figure \ref{exp:tSNE} (b) shows that the data distributions of different languages are mixed and overlap each other, which indicates that our model induces language-independent features and boosts the multilingual representation alignment. 

\section{Conclusion}
This paper proposes progressive code-switching, which fully mines multilingual knowledge to enhance zero-shot cross-lingual performance. We first adopt a word relevance score calculation method to measure the difficulty of the code-switching data. Then we generate suitable code-switching data controlled by the adoptable temperature. Finally, we introduce a scheduler to decide when to sample harder data for model training. Experimental results on the three zero-shot cross-lingual tasks covering ten languages exhibit the effectiveness and potential of our proposed method. 
\section*{Acknowledgements}
We thank Beijing Advanced Innovation Center for Big Data and Brain Computing for providing computation resources. We also thank the anonymous reviewers and the area chair for their insightful comments. This research was supported by the Shanxi Province Special Support for Science and Technology Cooperation and Exchange (202204041101020), Key Laboratory of Key Technologies of Major Comprehensive Guarantee of Food Safety for State Market Regulation (BJSJYKFKT202302), and the Zhejiang Provincial Natural Science Foundation of China under Grant (LGG22F020043).


\bibliographystyle{named}

\end{document}